# Older adults' acceptance of SARs: The link between anticipated and actual interaction

Maya Krakovski[1†], Oded Zafrani[1†], Galit Nimrod[2], Yael Edan[1]

**This study demonstrates how anticipated and actual interactions shape the QE of SARs among older adults. The study consisted of two parts: an online survey to explore the anticipated interaction through video viewing of a SAR and an acceptance study where older adults interacted with the robot. Both parts of this study were completed with the assistance of Gymmy, a robotic system that our lab developed for training older adults in physical and cognitive activities. Both study parts exhibited similar user responses, indicating that users' acceptance of SARs can be predicted by their anticipated interaction.**

***Index Terms*** **— Aging, human-robot interaction, older adults, quality evaluation, socially assistive robots, technology acceptance, technophobia, trust, user experience.**

## I. Introduction

The development of social assistive robots (SARs) has been considered a viable solution for the shortage of workers in many fields (e.g., healthcare, service, tourism), as well as a way to overcome social distance challenges, which was necessary during the COVID-19 pandemic. As the world's population ages rapidly, and the number of older adults is projected to increase significantly in the coming years [1], many SAR developments focus on this population. The growing interest in human-robot interaction (HRI) design has introduced new challenges in evaluating SARs. HRI studies are commonly conducted in two conditions: video-based demonstration of an HRI scenario which infers to anticipated interaction, or with a real robot which relates to actual interaction with SARs [1]. The latter is usually conducted in lab studies [2] or under real-life conditions [3], [4].

In anticipated interaction studies, it is possible to engage a larger number of participants, incorporate new ideas more easily, test initial assumptions, and have greater control over experimental conditions. In actual interaction studies, participants interact with robots in real-life environments, resulting in an evaluation that is more realistic to using SARs in daily life. Currently, most literature focuses on one type of interaction (anticipated or actual). We propose combining both types of interactions to obtain more meaningful observations and a comprehensive understanding of the reciprocity between humans and robots.

Assessing HRI by mixing quantitative and qualitative methods allows for a more accurate assessment [5]. Many of the evaluation methods in robotics and HRI are quantitative in nature since the field is led by engineers and computer scientists. As understanding that HRI should be a cross-disciplinary field involving researchers from social science grows, the value of qualitative evaluation is being recognized.

For successful acceptance and assimilation of SARs, it is necessary to understand the factors affecting their Quality Evaluations (QEs) by older adults. There are hedonic and pragmatic aspects to QE [6]. The hedonic aspects of users' experiences relate to their emotional inclinations towards the technology during the interaction, whereas the pragmatic aspects relate to the usability, usefulness, and utility of potential tasks [6]. A product's attractiveness is determined by both hedonic and pragmatic evaluations [7]. Positive QEs are necessary to promote the acceptance of SARs [8].

This study aims to demonstrate how the QE of SARs among older adults is shaped by anticipated and actual interaction. Accordingly, it was carried out in two parts: (a) an online survey with 384 participants to explore the anticipated interaction through video viewing of a SAR and (b) an acceptance study with 26 participants in which the older adults interacted with the robot. In both parts, we used "Gymmy," a robotic system developed in our lab for older adults' physical and cognitive training. The system design includes a humanoid mechanical-looking robot to demonstrate exercises, an RGB-Depth (RGB-D) camera to measure performance, and a touch screen and speakers to provide instructions and feedback. Older adults over 65 years participated in the studies.

## II. Anticipated Interaction

The effect of trust and technophobia on older adults' QE of SARs was examined through an online survey of 384 respondents [9]. Participants were invited to watch a three-minute video that presented Gymmy and its functions (www.youtube.com/watch?v=zQ4T1NhS25Q). After watching the video, they were asked to answer questions related to Gymmy.

The online survey results indicated that participants perceived Gymmy as providing more pragmatic value (Mean=1.46, SD=1.18) than a hedonic experience (Mean=0.63, SD=1.42) and trusted its performance-based functional capabilities more than its social aspects [2]. In addition, a positive association was found between trust and QE (r=0.43, $p<.01$) while a negative association was found between technophobia and QE (r=-0.56, $p<.01$). The relative impact of technophobia on older adults' QE of SARs was significantly more substantial than that of trust, and the robot-

*Research supported by the Ministry of Science grant agreement number 3-15713. Partial support was provided by Ben-Gurion University of the Negev through the Agricultural, Biological and Cognitive Robotics Initiative, the Marcus Endowment Fund, and the W. Gunther Plaut Chair in Manufacturing Engineering.

[1]Industrial Engineering & Management Department, Ben-Gurion University of the Negev, Beer Sheva, Israel.

[2]Department of Communication Studies, Ben-Gurion University of the Negev, Beer Sheva, Israel.

[†]These authors contributed equally to this work.

related technophobia constituted a most influential antecedent constraint to SARs use.

## III. Actual Interaction

Twenty-six older adults participated in the study, including an actual interaction with Gymmy in their home. The study examined the effect of participants' characteristics and training on the users' acceptance of the robot. The acceptance was measured by four components of the technology acceptance model [10]: perceived usefulness, ease of use, attitude, and intention to use.

The outcomes of the actual interaction revealed that the user's attitude toward interactions with robots had a significant association with their rating of perceived usefulness, ease of use, and intention to use the robotic trainer. In the pre-experiment questionnaire, most participants indicated they do not have negative feelings toward interaction with robots. The post-experiment questionnaire results showed that about half of the participants found the robot useful (Mean=3.61, SD=1.19), and about a third found it easy to use (Mean=4.15, SD=0.91). In addition, nearly half of the participants indicated a high intention to use the robotic system in the future (Mean=3.5, SD=1.26).

The results also showed that age group, education, and initial attitude towards robots influenced the acceptance of the robotic system. Overall, the actual interaction study showed that the system was robust and reliable, demonstrating that it is potentially helpful in motivating older adults to engage in physical and cognitive activities. Further insights from this study revealed that users' attitudes are influenced by their success during training. The fact that participants requested more explanatory feedback from the robotic exercise trainer and that the system's feedback architecture indicates success in exercises indicates that feedback plays a critical role in the robotic exercise trainer.

## IV. Comparison

Comparing the anticipated and the actual interaction studies suggests that the acceptance level can be predicted based on anticipated interaction. The online survey highlighted that the relative impact of technophobia is significantly more substantial than that of trust and that the pragmatic qualities of the robot and its convenience of use in particular are more crucial to its QE than the emotional aspects of use.

Similar to the online survey results, a link between the participants' attitudes toward robots and their acceptance of the robotic trainer was found in the experiment with the actual robot interaction. As most of the participants did not indicate any negative attitude towards robots before the experience with the robot, the users indicated acceptance after training with the robotic trainer. The participants' initial attitudes towards robots were found to influence their ratings of the robotic trainer's usefulness, ease of use, and intention to use the robotic trainer in the future.

The similar responses in both study parts imply that users' acceptance of SARs can be predicted by their anticipated interaction. Since often experiments with SARs, especially those involving older adults, require significant resources and have less control over simulations, video studies should become a vital observation [11]. As an initial evaluation, it is possible to experiment with anticipated interaction between robots and users to evaluate the robots' functionality and user demands. However, this should be only for preliminary user design concepts. To ensure an accurate evaluation, investigations should also include real-world and actual interactions with robots since users may experience differences once they encounter a real robot. Researchers have already shown that older adults prefer a robotic trainer that is physically embodied rather than a simulator [12]. Since lab studies are limited as the interaction with the robot is partially artificial due to the controlled environment. actual operation in real-life conditions is important. This is especially important with nonprofessional users, as results from actual interaction experiments consider the challenges of daily life reality. Additionally, studies should include longitudinal methods to assess assimilation and QE [9].

The results further revealed the importance of customizing the system to meet the needs of different users, as well as the importance of feedback. Moreover, the findings highlight the importance of investigating technophobia in HRI studies. Implementing robotics technology in later life strongly depends on reducing older adults' sense of technophobia. Future research must find ways to address technophobia at the early stages of technology design.

## Appendix

User Experience Questionnaire (UEQ) [13].

The following items present ways in which people may describe social robots. Regarding each pair of items, please mark your overall impression with the social robot that you have used in the past weeks.

| | -3 | -2 | -1 | 0 | +1 | +2 | +3 | |
|---|---|---|---|---|---|---|---|---|
| annoying | ○ | ○ | ○ | ○ | ○ | ○ | ○ | enjoyable |
| not understandable | ○ | ○ | ○ | ○ | ○ | ○ | ○ | understandable |
| creative | ○ | ○ | ○ | ○ | ○ | ○ | ○ | dull |
| easy to learn | ○ | ○ | ○ | ○ | ○ | ○ | ○ | difficult to learn |
| valuable | ○ | ○ | ○ | ○ | ○ | ○ | ○ | inferior |
| boring | ○ | ○ | ○ | ○ | ○ | ○ | ○ | exciting |
| not interesting | ○ | ○ | ○ | ○ | ○ | ○ | ○ | interesting |
| unpredictable | ○ | ○ | ○ | ○ | ○ | ○ | ○ | predictable |
| fast | ○ | ○ | ○ | ○ | ○ | ○ | ○ | slow |
| inventive | ○ | ○ | ○ | ○ | ○ | ○ | ○ | conventional |
| obstructive | ○ | ○ | ○ | ○ | ○ | ○ | ○ | supportive |
| good | ○ | ○ | ○ | ○ | ○ | ○ | ○ | bad |
| complicated | ○ | ○ | ○ | ○ | ○ | ○ | ○ | easy |
| unlikeable | ○ | ○ | ○ | ○ | ○ | ○ | ○ | pleasing |
| usual | ○ | ○ | ○ | ○ | ○ | ○ | ○ | leading edge |
| unpleasant | ○ | ○ | ○ | ○ | ○ | ○ | ○ | pleasant |
| secure | ○ | ○ | ○ | ○ | ○ | ○ | ○ | not secure |
| motivating | ○ | ○ | ○ | ○ | ○ | ○ | ○ | demotivating |
| meets expectations | ○ | ○ | ○ | ○ | ○ | ○ | ○ | does not meet expectations |
| inefficient | ○ | ○ | ○ | ○ | ○ | ○ | ○ | efficient |
| clear | ○ | ○ | ○ | ○ | ○ | ○ | ○ | confusing |
| impractical | ○ | ○ | ○ | ○ | ○ | ○ | ○ | practical |
| organized | ○ | ○ | ○ | ○ | ○ | ○ | ○ | cluttered |
| attractive | ○ | ○ | ○ | ○ | ○ | ○ | ○ | unattractive |
| friendly | ○ | ○ | ○ | ○ | ○ | ○ | ○ | unfriendly |
| conservative | ○ | ○ | ○ | ○ | ○ | ○ | ○ | innovative |